\documentclass[runningheads]{llncs}
\usepackage{graphicx}
\usepackage{booktabs}
\usepackage{tabularx}
\usepackage{hyperref}
\usepackage{paralist}
\usepackage{tabulary}
\usepackage{ragged2e}
\usepackage{wrapfig}
\usepackage{cite}
\usepackage{comment}
\usepackage{caption}
\usepackage{amsmath}
\usepackage{color,soul}
\usepackage{multirow}
\usepackage{tabularray}
\usepackage{float}

\newcolumntype{Y}{>{\centering\arraybackslash}X}
\newcommand{\mypar}[1]{\vspace{0.1em}\noindent\textbf{#1.}}

\begin{document}
\title{Large Language Models can accomplish\\ Business Process Management Tasks}
\titlerunning{Large Language Models in Business Process Management}
%
%\titlerunning{A Reference Data Model for UI Logs}
\author{Michael Grohs$^*$ \and Luka Abb$^*$ \and Nourhan Elsayed$^*$ \and Jana-Rebecca Rehse}
\authorrunning{Grohs et al.}
\institute{University of Mannheim \\
\email{\{michael.grohs,luka.abb,elsayed,rehse\}@uni-mannheim.de}}
\maketitle              % typeset the header of the contribution
\def\thefootnote{*}\footnotetext{Equal contribution}\def\thefootnote{\arabic{footnote}}
\begin{abstract}
Business Process Management (BPM) aims to improve organizational activities and their outcomes by managing the underlying processes. To achieve this, it is often necessary to consider information from various sources, including unstructured textual documents. Therefore, researchers have developed several BPM-specific solutions that extract information from textual documents using Natural Language Processing techniques. 
These solutions are specific to their respective tasks and cannot accomplish multiple process-related problems as a general-purpose instrument. However, in light of the recent emergence of Large Language Models (LLMs) with remarkable reasoning capabilities, such a general-purpose instrument with multiple applications now appears attainable. 
In this paper, we illustrate how LLMs can accomplish text-related BPM tasks by applying a specific LLM to three exemplary tasks: mining imperative process models from textual descriptions, mining declarative process models from textual descriptions, and assessing the suitability of process tasks from textual descriptions for robotic process automation. We show that, without extensive configuration or prompt engineering, LLMs perform comparably to or better than existing solutions and discuss implications for future BPM research as well as practical usage.

\end{abstract}
\section{Introduction}

%%1st Paragraph: process information often in text; information abstraction necessary; automated approaches for different tasks proposed

%%2nd Paragraph: LLMs developed and their capabilities + how they can be used to replace

The objective of Business Process Management (BPM) is to understand and supervise the execution of work within an organization. This ensures consistent outcomes and allows for the identification of improvement opportunities \cite{bpm_book}. To accomplish this, BPM researchers and practitioners make use of diverse sources of information pertaining to business processes. These sources range from well-structured process models and event logs to unstructured textual documents \cite{Aa2018}. In the past decade, BPM researchers have increasingly turned to Natural Language Processing (NLP) techniques to automatically extract process-related information from the abundant textual data found in real-world organizations.
%The recent significant advancements in the field of NLP have further fueled the already substantial interest in this area of BPM.

Many existing approaches utilize textual data for a wide range of BPM tasks. Examples of such tasks include the mining of imperative or declarative process models from textual process descriptions \cite{friedrich2011process,van2019extracting}, process redesign for classifying end-user feedback  \cite{mustansir2022towards}, identifying suitable tasks for robotic process automation (RPA) in textual process descriptions \cite{leopold2018identifying}, assessing process complexity based on textual data \cite{rizun2021assessing}, or extracting semantic process information from natural language \cite{rebmann2021extracting}.
Although a few approaches also incorporate machine learning methods, the majority rely on extensive rule sets.
%\cite{mustansir2022towards,leopold2018identifying}.
%\hl{citations}.

Each existing approach is designed for a specific purpose, meaning that it can only be applied to one specific task. A versatile general-purpose model that comprehends process-related text and seamlessly integrates it into various BPM tasks does not yet exist. However, the recent emergence of pre-trained Large Language Models (LLMs), which have demonstrated remarkable reasoning abilities across diverse domains and tasks \cite{teubner2023welcome}, offers promising prospects for developing such a system.
%LLMs are neural networks for language generation, which are built on the transformer architecture and trained on huge amounts of data.
Already, multiple research groups are actively exploring the potential of these models in the BPM field, for example by analyzing which opportunities and challenges LLMs pose for the individual stages of the BPM lifecycle \cite{vidgof2023large}, how LLMs input should look like such that the output supports BPM \cite{busch2023just}, or whether conversational process modeling is possible \cite{klievtsova2023conversational}.
%These recent publications are circulating pre-prints and not peer-reviewed yet. 
%Evidently, the BPM field is starting to explore the potentials of LLMs for research and practice.

These recent publications and pre-prints mostly illustrate the potential and difficulties of LLMs on a high level, but they do not showcase concrete applications. In this paper, we take a more application-oriented approach by investigating whether an LLM can accomplish three text-related BPM tasks: 
(1) mining imperative process models from textual descriptions,
(2) mining declarative process models from textual descriptions,
and (3) assessing the suitability of process tasks for RPA from textual descriptions. 
We selected these tasks because they are practically relevant and have previously been addressed in research.
We evaluate how well the LLM can perform these tasks by benchmarking them against existing approaches that were specifically developed for the respective task. Based on the results, we discuss implications for future research in the field of BPM and illustrate how LLMs can support practitioners in their daily work.

The paper is structured as follows: In Sect. \ref{sec:approach}, we introduce the general solution approach that we followed for all three tasks. The task-specific applications and results are described in Sect. \ref{sec:ltl}, Sect. \ref{sec:bpmn}, and Sect. \ref{sec:rpa}, respectively. Section \ref{sec:disc} discusses the future usage of LLMs in practice as well as implications for future research, before we conclude the paper in Sect. \ref{sec:concl}.

\begin{figure}[htb]
%\vspace{-1.5em}
\centerline{\includegraphics[trim=50 230 50 235,clip,width=0.95\textwidth]{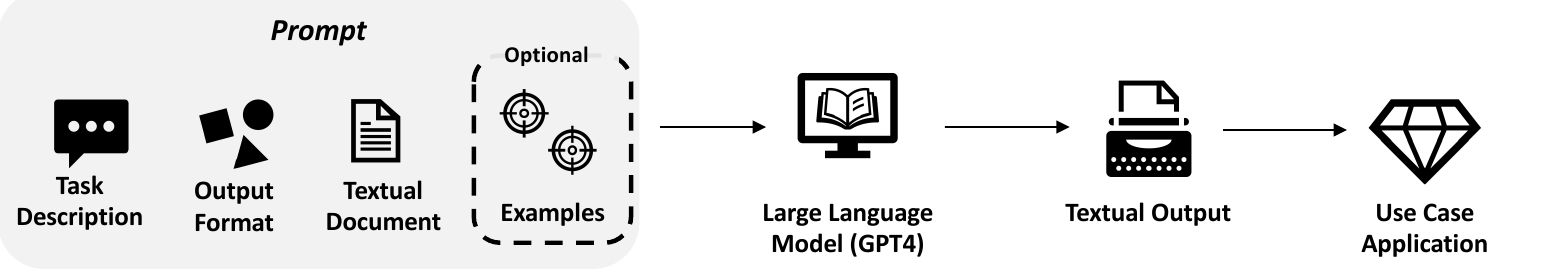}}
%\vspace{-1em}
\caption{Overview of our Approach}\label{fig:approach}
\end{figure} 
%\vspace{-3em}

\section{Approach} \label{sec:approach}

In this paper, we illustrate how LLMs can be utilized for three BPM tasks that require textual documents as input. For all tasks, we follow the same approach, illustrated in Fig.~\ref{fig:approach}. We start by assembling a prompt with the following parts:

\begin{enumerate}
\item A general description of the BPM task that is to be accomplished. 
\item A specification of a particular output format that the LLM should adhere to. This ensures that the generated text output has a certain level of consistency and that results are sufficiently standardized so that they can be further processed by, for example, parsing algorithms. 
\item The natural language text that we want to abstract information from, e.g., a textual process description
\item Optionally, if suitable for a given task, few input-output pairs as examples
\end{enumerate}

\noindent The complete prompt is then entered into the current state-of-the-art instruction-following LLM, ChatGPT with GPT4 backend \cite{openai2023gpt4} (henceforth referred to as GPT4). The textual output of GPT4 (i.e., the response to the prompt) is then evaluated with respect to its utility in solving the respective task and benchmarked against an existing approach. All parts of the prompt have not been specifically engineered but rather included such that the output is actually solving the tasks. The prompts were not optimized with respect to any metric.

In all applications, we provide the model with several prompts in order to check input robustness (i.e., how prompts from different authors influence the results) and output robustness (i.e., how the results change for different tries of the same prompt). By this, we aim to analyze whether GPT4 is able to accomplish specific tasks sufficiently well to be used by a diverse group of people and whether the results remain consistent despite the inherent randomness of the model output. For each task, we start with an ´´original´´ prompt written by one of the authors of this paper and enter this prompt three times (Tries 1 to 3). Two more prompts are then created by two other authors, who are given a general description of the task to be accomplished and the exact output format that they should specify, but who have not seen the original prompt. Finally, where appropriate, we also enter the original prompt without examples to evaluate the effect those have on the result. Each prompt is entered in a separate conversation window in the GPT4 web interface so that the model cannot draw on previous prompts as context.

All prompts, responses, and detailed evaluation results are available online\footnote{\url{https://gitlab.uni-mannheim.de/jpmac/llms-in-bpm}}.
%Also, we thus analyze how appropriate prompts should look for the task to be successful. 
%and is based on the abstracted information.

%\section{Exemplary use cases}

\section{Mining BPMN process models from natural language descriptions} \label{sec:bpmn}

\subsection{Motivation}
Process models are the predominant tool for representing organizational activities and are often the starting point for process analyses \cite{van2019extracting}. 
%Several modelling languages exists to display processes in an imperative manner (i.e., a process model that shows which behavior is allowed) or in a declarative manner (i.e., a process model that shows which behavior is forbidden) \cite{van2019extracting}. 
Constructing such models requires knowledge of the process and proficiency in the creation of formal models \cite{friedrich2011process}. However, the actors with process knowledge commonly are not experienced process modelers \cite{friedrich2011process}. Therefore, modeling procedures can be very time-consuming and error-prone \cite{reijers2003product}. This holds true even though detailed textual descriptions of process requirements are often available in the form of policies, guidelines, or e-mail conversations, which can be considered relevant sources of information \cite{friedrich2011process}. Approaches that extract process models from natural language can speed up the modeling and also enable managers to frequently update their process models without requiring extensive modeling experience.

A rule-based approach to extract Business Process Model and Notation (BPMN) process models from textual process descriptions was first proposed in \cite{friedrich2011process}. This remains the only generally-applicable, end-to-end technique able to produce a full imperative process model from text input, though several other publications with a more narrow scope or a focus on mining partial models exist (see \cite{Bellan2020} for a short review). There are also papers that investigate the ability of LLMs to extract process entities and relations from textual descriptions \cite{klievtsova2023conversational,Bellan2022}. Though their approaches have some similarities to ours, neither ends up producing an actual process model from the text.

\subsection{Evaluation}
Following Fig.~\ref{fig:approach}, we ask GPT4 to create a BPMN model for a process described in text. At the time of writing, the web interface version of GPT4 has a token output limit that prevents it from generating sequences at the length that would be required to generate BPMN models as XML files. We, therefore, prompt it to produce a model in a pre-specified intermediary notation as an output format that includes the main elements of BPMN and is straightforward to parse into a proper model representation. The template we provided in the prompt represents task nodes as natural language words, arcs between model elements as arrows (\texttt{-$>$}), and exclusive and parallel gateways as \texttt{XOR} and \texttt{AND}, respectively. We also specify that outgoing arcs of exclusive gateways can be labeled to represent decision criteria, e.g., \texttt{XOR (Proposal accepted) -$>$ \ Task1}. Finally, we ask the model to provide an actor-to-activity mapping that can be used to construct lanes, in the format \texttt{actor: [activity1, ...]}. Other elements (e.g., messages) are not included. 
We also do not provide example pairs of text and corresponding full or partial models to the LLM to avoid bias towards a certain modeling style.

\begin{figure}[htb]
\centering
\includegraphics[width=0.35\textwidth]{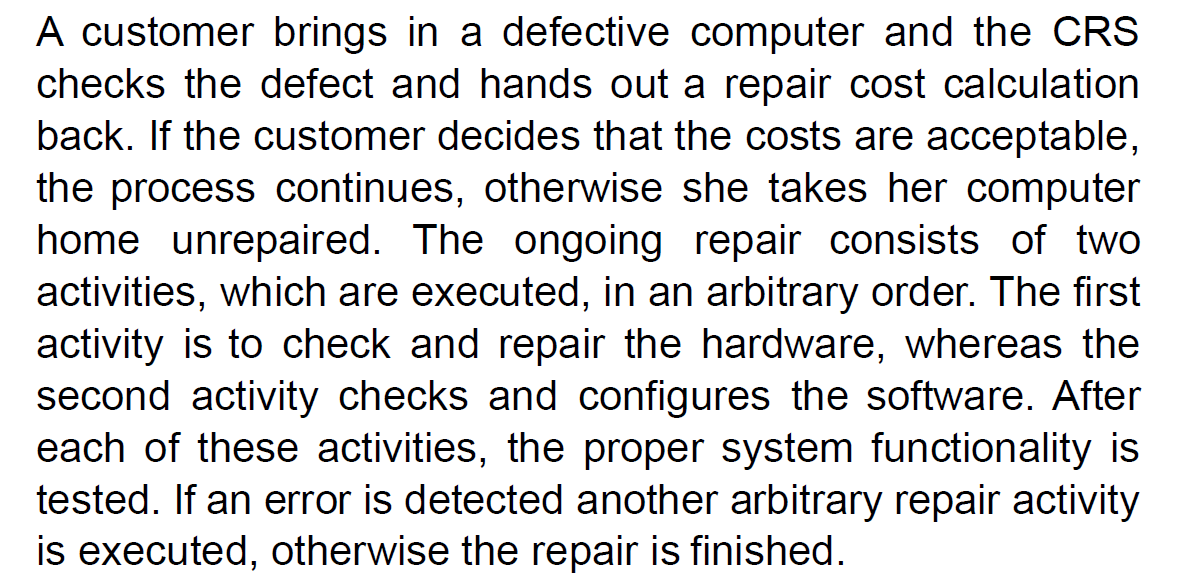}
\includegraphics[width=0.38\textwidth]{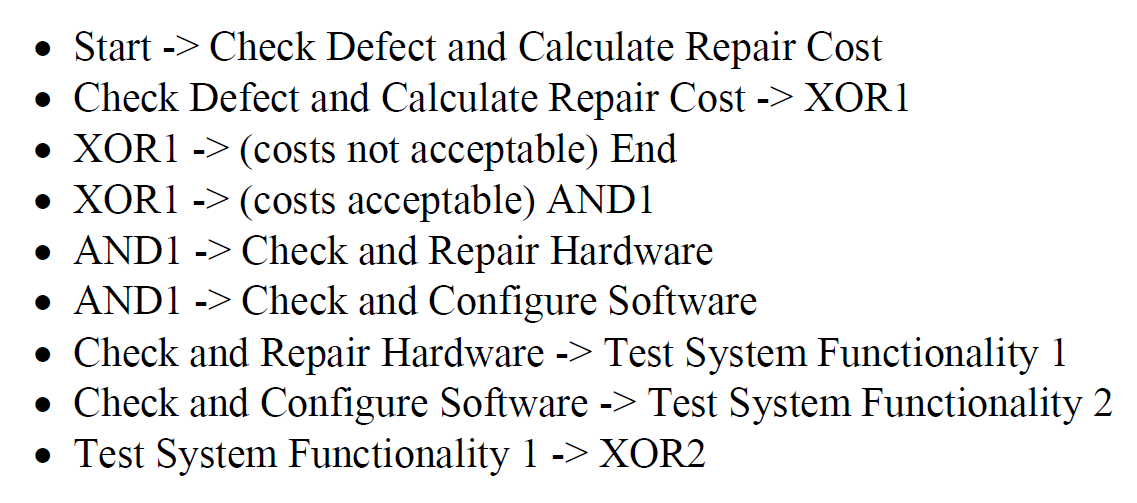}
\includegraphics[width=0.8
\textwidth]{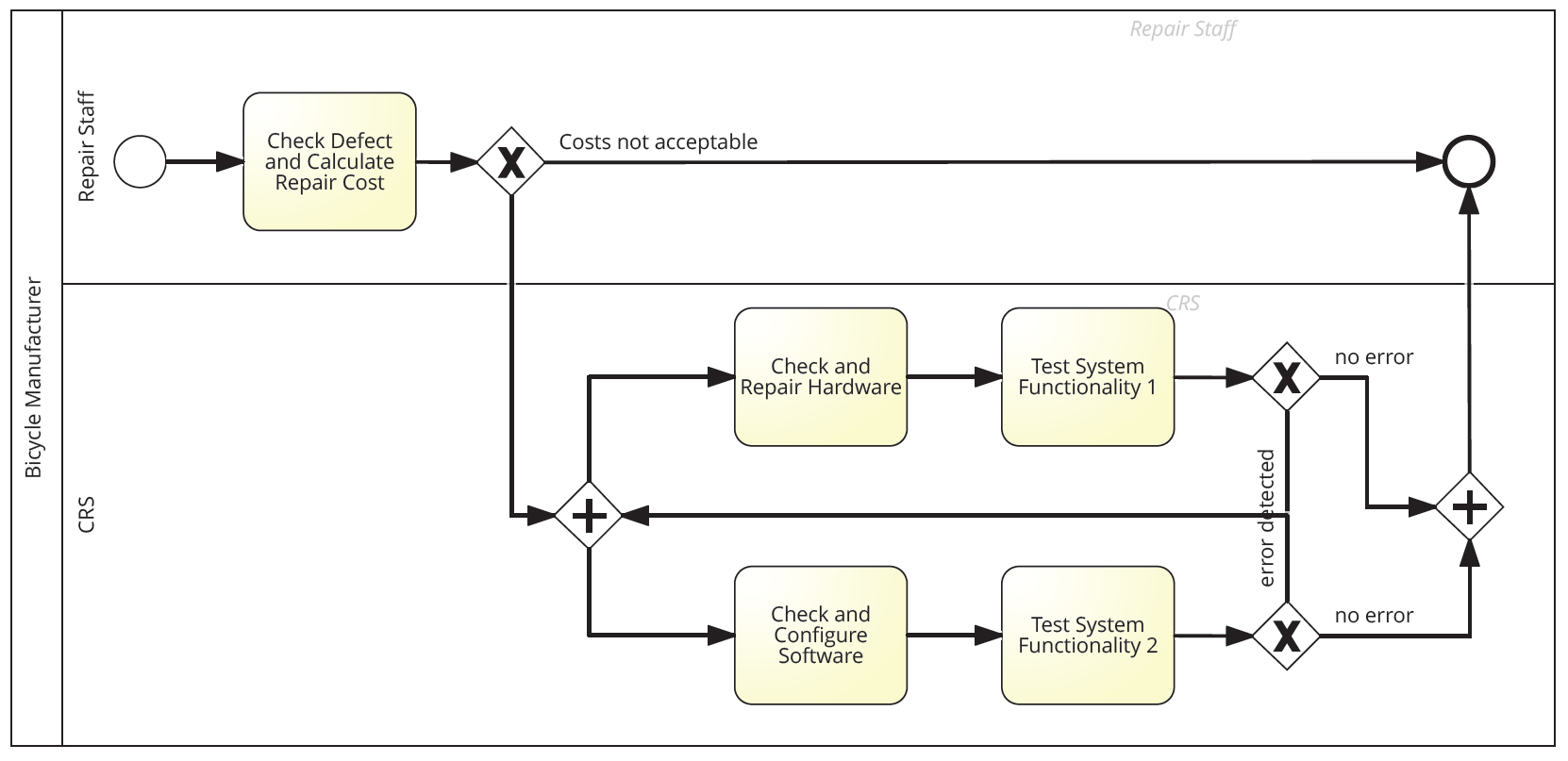}
\caption{Example of a textual process description (top left), an excerpt of the generated LLM response (top right, from Prompt 1 Try 1), and visualization of the corresponding BPMN diagram (bottom).}
\label{fig:bpmn_example}
%\vspace{-1.5em}
\end{figure}

Figure \ref{fig:bpmn_example} shows an example of a textual description of a computer repair process, an excerpt of the response that GPT4 gave when presented with this description, and a visualization of the derived BPMN model. The generated model accurately represents the process described in the text. It could, however, be made slightly simpler by combining the two separate \emph{Test System Functionality} activities and the subsequent exclusive gateways into one each.

For our evaluation, we use six process descriptions from \cite{friedrich2} (1.1 - 1.4, 2.1, and 2.2). We selected these with the goal of applying our technique to a mix of short and simple as well as longer and more sophisticated textual descriptions. As ground truth, we use the annotations provided for these descriptions in the PET dataset \cite{PET}. Specifically, we evaluate the output of the LLM with regard to how many of the relations described in the textual description are correctly identified (i.e., recall). Note that this allows us to simultaneously evaluate how many entities (task names and actors) are correctly identified since a relation that involves an unknown entity will be counted as not identified. We do not evaluate the models with regard to how many superfluous entities or relations they produce (i.e., precision) as that would raises several conceptual questions that require answers (e.g., how to treat a task that is correctly identified but in the wrong position), which would go beyond the intended scope of this paper.

We further restrict our evaluation to \emph{flow} and \emph{actor performer} relations, i.e., those that are present in the intermediary notation we provide in the prompts. Since the ground truth annotation applies only to the textual descriptions, we manually establish a mapping between the entities identified in the dataset and the ones produced by GPT4. In some cases, the relations produced by the LLM do not exactly match the ground truth (e.g., \texttt{Write Report} and \texttt{Send Report} are combined to \texttt{Write and Send Report}). For these, we follow the same approach as \cite{Bellan2022}, i.e., we evaluate them on a case-by-case basis and count them as correct if they are semantically correct. As a benchmark, we use the process models produced by \cite{friedrich2}, applying the same evaluation criteria as described above.

\begin{table}[h]
%\vspace{-2em}
\caption{Recall for the Text-to-BPMN Task}
\label{tab:eva_BPMN}
\centering 
\begin{tabularx}{\textwidth}{|c|X||>{\centering\arraybackslash}m{1cm}>{\centering\arraybackslash}m{1cm}>{\centering\arraybackslash}m{1cm}>
{\centering\arraybackslash}m{1cm}>{\centering\arraybackslash}m{1cm}>
{\centering\arraybackslash}m{1cm}|>{\centering\arraybackslash}m{1.6cm}|}
\hline
\multicolumn{1}{|l}{} & \multicolumn{1}{l||}{} & Text 1.1 & Text 1.2 & Text 1.3 & Text 1.4 & Text 2.1 & Text 2.2 & Overall \\ \hline \hline
\parbox[t]{3mm}{\multirow{3}{*}{\rotatebox[origin=c]{90}{\parbox[c]{.8cm}{\centering OR}}}} & \shortstack[l]{Prompt 1 Try 1} & 0.42 & \textbf{0.58} & 0.46 & 0.50 & 0.57 & 0.45 & 0.50 \\ \cline{2-9} 
 & \shortstack[l]{Prompt 1 Try 2}  & \textbf{0.54} & \textbf{0.58} & 0.38 & \textbf{0.70} & \textbf{0.61} & 0.42 & 0.54 \\ \cline{2-9}
 & \shortstack[l]{Prompt 1 Try 3}  & \textbf{0.54} & \textbf{0.58} & 0.50 & 0.60 & 0.53 & 0.53 & 0.54 \\ \hline
\parbox[t]{3mm}{\multirow{2}{*}{\rotatebox[origin=c]{90}{\parbox[c]{.8cm}{\centering IR}}}} & \shortstack[l]{Other Author (1)} &  \textbf{0.54} & 0.47 & 0.54 & 0.50 & 0.47 & 0.34 & 0.48 \\ \cline{2-9}
& \shortstack[l]{Other Author (2)} &  0.46 & 0.42 & 0.35 & 0.47 & 0.43 & 0.39 & 0.42 \\ \hline
%\multicolumn{2}{|l||}{Aggregated}  & 0.49 & 0.53 & 0.45 & 0.56 & 0.52 & 0.43 & 0.50 \\ \hline
\multicolumn{2}{|l||}{Benchmark \cite{van2019extracting}} & \textbf{0.54} & 0.47 & \textbf{0.58} & 0.55 & 0.55 & \textbf{0.66} & \textbf{0.56} \\ \hline 
\end{tabularx}
\vspace{-.5em}
\end{table}

The results of our evaluation are shown in Tab. \ref{tab:eva_BPMN}, subdivided by the evaluation of output robustness (OR) and input robustness (IR). Overall, regarding the proportion of relations (and entities) that are correctly extracted from the textual process description, the models generated by GPT4 are comparable to the ones produced by \cite{friedrich2011process}. Note that the absolute numbers reported should be interpreted with caution, because the PET ground truth is very fine-granular and we weigh all relation types equally, so that, for example, a single missing exclusive gateway (with the gateway itself, two decision criteria on the outgoing arcs, and two subsequent activities) would be counted as five non-identified relations. Consequently, a recall value of 0.5 should not be understood to indicate that the model only includes half of the relevant process behavior described in the text.
%The prompts written by the two other authors also resulted in responses that did not include arcs between task nodes, which significantly draws down their performance. This is simply a prompting issue, but we decided to include results as they are rather than re-doing these in order to draw attention to the discrepancy in output quality that can result from inadequate prompting. 
Furthermore, 
%it should be taken into account that 
the models generated by GPT4 are very precise in the sense that they tend to include a minimal (often insufficient) set of tasks, whereas the rule-based approach of \cite{friedrich2011process} tends to produce models with several superfluent activities (e.g., \texttt{Begin Process} following a start event). Since our evaluation does not include a notion of false positive relations, it could be argued that we somewhat underestimate the quality of the LLM output relative to the benchmark.

Overall, an LLM-based text-to-BPMN technique produces reasonably good results. The model also produces consistent answers in the same intermediary notation when provided with the exact same description of the target template, so parsing its output into XML is possible. With prompt fine-tuning, and especially with subsequent prompting that asks the model to fix common issues, it is not unfeasible to create a reliable text-to-BPMN pipeline based on an LLM.

\section{Mining declarative process models from natural language descriptions} \label{sec:ltl}

\subsection{Motivation}

Not all business processes can be adequately captured by imperative modeling notations such as BPMN. For instance, knowledge-intensive processes have execution orders that cannot always  be fully specified in advance \cite{van2019extracting}. These are better modeled using declarative process models, i.e., a set of formal constraints that do not rely on an explicit definition of allowed behavior \cite{burattin2016conformance}. They provide a flexible way of modeling processes, especially suitable in complex settings \cite{burattin2016conformance}.

An approach that extracts declarative process models from natural language has been proposed in \cite{van2019extracting}. It uses the common declarative modeling language \emph{Declare}, which is based on constraint templates grounded in Linear Temporal Logic (LTL) \cite{burattin2016conformance}. By applying rule-based NLP techniques to sentences, the approach in \cite{van2019extracting} generates declarative constraints for five LTL templates: precedence, response, succession, initialization (init), and end. 
%The first three prescribe the relative ordering of two activities: 
\textit{Precedence(A, B)} (formal as \texttt{NOT(B) U A}) means that activity B should only occur after activity A. \textit{Response(A,B)} (formal as \texttt{A -$>$ B}) means that B must follow whenever A occurs. \textit{Succession(A,B)} is the combination of \textit{Precedence(A,B)} and \textit{Response(A,B)}. \textit{Init(A)} (formal as \texttt{START -$>$ A}) prescribes that all process instances must start with A and \textit{End(A)} (formal as \texttt{END -$>$ A}) indicates that they must end with A. 

\subsection{Evaluation}
In our experiment, we recreate the set-up from \cite{van2019extracting}, applying GPT4 on the same five LTL templates and 104 test sentences. Following Fig.~\ref{fig:approach}, we create a prompt that asks GPT4 to create LTL formulas in the form of precedence, response, succession, init, and end. For each template, we provide the output format and an example. As a result, the LLM outputs one or more discovered constraints in the format prescribed by the prompt, as shown in the exemplary excerpt of the output in Tab.~\ref{tab:ex_ltl}. This output can then be compiled and translated into declarative modeling languages like \emph{Declare}.

\addtolength{\tabcolsep}{5pt}
\begin{table}[b]
\vspace{-2em}
    \centering
    \caption{Exemplary Output of GPT4 for the Text-to-LTL Task}
    \label{tab:ex_ltl}
    \begin{tabularx}{\textwidth}{XXl}
    \toprule
         \textbf{Sentence Input} & \textbf{GPT4 Output} & \textbf{LTL-Template} \\ \midrule
         A claim   should be created before it can be approved. & NOT(approve claim) U create claim & Precedence \\ \midrule
        The process begins with the booking of the ticket. & START -$>$  book ticket & Init \\ \midrule
        Every provided laundry service must be billed. & provide laundry service -$>$ F(bill) & Response \\ \bottomrule
    \end{tabularx}
%\vspace{-1.5em}
\end{table}
\addtolength{\tabcolsep}{-5pt}

%As introduced in Sec.~\ref{sec:approach}, we test the output robustness and input robustness, each with three different prompts. 
In addition to the three identical prompts for output robustness, we use two other formulations by different authors and, as we use examples in the original prompt, also one prompt without examples for input robustness.
%By this, we aim to analyze whether GPT4 is able to generate declarative process models sufficiently well to be used by a diverse group of humans. 
%We also want to analyze how appropriate prompts should look for the task to be successful. 
%The used prompts as well as the strings returned by GPT4 can be found in the repository referenced in Sec.~\ref{sec:bpmn}. 
%The labels have been set in accordance with the authors of \cite{van2019extracting}. 
Table \ref{tab:eva_LTL} displays precision (Prec.), recall (Rec.), and F1-score (F1) as used by \cite{van2019extracting} for each of the five LTL templates of the six different prompts compared to the benchmark.\footnote{The corresponding confusion matrices can be found in our repository.} We only consider syntactically correct classifications as true positives.
%Note that we display only on the evaluation of the templates and not the same table used in \cite{van2019extracting} portraying negation, single- and multi-constraint sentences,
%due to space restrictions
%but rather summarize all this in the correct recognition of the five templates.

\begin{table}[h]
\caption{Precision, Recall, and F1-Score for the Text-to-LTL Task}
\label{tab:eva_LTL}
\centering
\begin{tabularx}{\textwidth}{|c|X|l||>{\centering\arraybackslash}m{1.6cm}>{\centering\arraybackslash}m{1.6cm}>{\centering\arraybackslash}m{1.5cm}>{\centering\arraybackslash}m{1cm}>{\centering\arraybackslash}m{1cm}|>{\centering\arraybackslash}m{1.4cm}|}
\hline
\multicolumn{1}{|l}{} & \multicolumn{1}{l}{} &  & Precedence & Response & Succession & Init & End & Overall \\ \hline \hline
\parbox[t]{2mm}{\multirow{9}{*}{\rotatebox[origin=c]{90}{Output Robustness}}} & \multirow{3}{*}{\shortstack[l]{Prompt 1\\ Try 1}} & Prec. & \textbf{0.96} & 0.68 & \textbf{1.00} & \textbf{1.00} & \textbf{1.00} & 0.84 \\
 &  & Rec. & 0.53 & \textbf{0.96} & \textbf{1.00} & 0.82 & 0.17 & 0.76 \\
 &  & F1 & 0.68 & \textbf{0.80} & \textbf{1.00} & 0.90 & 0.29 & 0.79 \\ \cline{2-9}
 & \multirow{3}{*}{\shortstack[l]{Prompt 1\\ Try 2}} & Prec. & 0.91 & 0.65 & \textbf{1.00} & \textbf{1.00} & \textbf{1.00} & 0.80 \\
 &  & Rec. & 0.57 & 0.75 & 0.33 & \textbf{1.00} & 0.50 & 0.68 \\
 &  & F1 & 0.70 & 0.70 & 0.50 & \textbf{1.00} & 0.67 & 0.74 \\ \cline{2-9}
 & \multirow{3}{*}{\shortstack[l]{Prompt 1\\ Try 3}} & Prec. & 0.94 & 0.68 & \textbf{1.00} & \textbf{1.00} & \textbf{1.00} & 0.83 \\
 &  & Rec. & 0.61 & 0.88 & 0.25 & \textbf{1.00} & 0.60 & 0.76 \\
 &  & F1 & 0.74 & 0.77 & 0.40 & \textbf{1.00} & 0.75 & 0.79 \\ \hline
\parbox[t]{2mm}{\multirow{9}{*}{\rotatebox[origin=c]{90}{Input Robustness}}} & \multirow{3}{*}{No Examples} & Prec. & 0.57 & 0.51 & 0.33 & 0.83 & \textbf{1.00} & 0.58 \\
& & Rec. & 0.08 & 0.79 & 0.67 & 0.91 & 0.50 & 0.49 \\
& & F1 & 0.14 & 0.62 & 0.44 & 0.87 & 0.67 & 0.53 \\ \cline{2-9}
& \multirow{3}{*}{\shortstack[l]{Other\\ Author (1)}} & Prec. & 0.94 & 0.72 & \textbf{1.00} & \textbf{1.00} & \textbf{1.00} & \textbf{0.86} \\
& & Rec. & 0.65 & 0.88 & \textbf{1.00} & 0.82 & 0.67 & \textbf{0.77} \\
& & F1 & \textbf{0.77} & 0.79 & \textbf{1.00} & 0.90 & 0.80 & \textbf{0.81} \\ \cline{2-9}
& \multirow{3}{*}{\shortstack[l]{Other\\ Author (2)}} & Prec. & 0.91 & 0.71 & 0.75 & 0.90 & \textbf{1.00} & 0.83 \\
& & Rec. & 0.61 & 0.77 & \textbf{1.00} & 0.82 & 0.83 & 0.72 \\
& & F1 & 0.73 & 0.74 & 0.86 & 0.86 & \textbf{0.91} & 0.77 \\ \hline
\multicolumn{2}{|c|}{\multirow{3}{*}{Benchmark \cite{van2019extracting}}} & Prec. & 0.78 & \textbf{0.8} & 0.68 & 0.75 & 0.88 & 0.77 \\
\multicolumn{2}{|c|}{} & Rec. & \textbf{0.71} & 0.77 & 0.68 & 0.82 & \textbf{0.88} & 0.72 \\
\multicolumn{2}{|c|}{} & F1 & 0.74 & 0.75 & 0.68 & 0.78 & 0.88 & 0.74 \\ \hline 
\end{tabularx}
%\vspace{-3em}
\end{table}

Except for the response template, GPT4 outperforms the benchmark and has a high precision value of close to 1. Further, we see that precision does not vary significantly with respect to output robustness for all LTL templates. With respect to recall, we see lower values for precedence. This is because many precedence constraints are misclassified as a response, which also explains the lower precision for this template. For succession and end, we see a high variation in the recall. This is due to a few constraints of these types in the 104 sentences, meaning that few misclassifications have a high impact. 
With respect to input robustness, the evaluation metrics are worse if no examples for the LTL templates are provided. This is especially visible for the precedence template. In contrast to that, different formulations from other authors do not have a significant impact on the metrics. Rather, stability across different prompts is visible.

The F1-score shows that all prompts with examples for the LTL templates yield equal or higher scores than the benchmark.
This illustrates that GPT4 outperforms the specific approach from \cite{van2019extracting} if it is provided with proper examples. 
This is an important finding as it indicates that prompts yield different results based on their fit to the task. Further, for tasks like this with short input text to be classified and a few classification targets, we recommend that the prompt should include examples. 
%Different prompt engineers do not affect this, indicating that GPT4 is robust to different formulations of the same task. Also, it yields comparable results for similar prompts, showing that the non-determinism of LLMs does not change the fact that they can be used for this BPM-specific task out-of-the-box. 
It should be noted that other prompts for example with additional information or the repetition of instructions could yield even better results. Further, the output of GPT4 has to be parsed into declarative process models using for example \emph{Declare} to allow complete usage. This is possible in an automatic manner given the consistent output format for all 104 sentences.

\section{Assessing RPA suitability of process tasks from natural language descriptions} \label{sec:rpa}

\subsection{Motivation}
RPA is a technology that aims to automate routine and repetitive tasks in business environments. To do so, software robots that work on the user interface of software systems are developed to perform these tasks the same way human actors would do, thus increasing operational efficiency \cite{reddy2019study}. 
 
%The RPA agent is deployed on user interfaces (UIs) to perform interactions with different interfaces of different software systems and applications \cite{torkhani2019collaborative}. Accordingly, this technology is suitable for tasks performed on Information Systems which will achieve higher operational efficiency \cite{dey2019robotic}. 

Various process information can be used to identify tasks that are suitable for RPA. This includes textual process descriptions, which are commonly used to document processes \cite{Aa2018}. The approach proposed in \cite{leopold2018identifying} identifies suitable tasks for RPA  by measuring the degree of the automation of process tasks using supervised machine learning techniques from the textual descriptions of business processes. 
% Organizations might have a lot of documents containing useful information about the business processes, their description, structure, or how the processes changed over time.
From this textual data, the approach classifies the process tasks into manual, automated, or user tasks. Manual tasks are the tasks performed by a human actor without any use of an information system, user tasks consist of humans interacting with an information system, and automated tasks are performed automatically on an information system without any human involvement. 
Tasks classified as user tasks are suitable RPA candidate tasks as they can be automated by replicating human interactions by means of RPA agents. This increases the efficiency of identifying suitable RPA tasks in comparison to a manual analysis that takes a long time and effort, especially
if there exists a large number of such documents or a large number of processes to be analyzed \cite{leopold2018identifying}.
%The classification helps to identify suitable RPA candidate tasks, which are user tasks, and automates the analysis of such process documents instead of the manual analysis that takes a long time and effort, especially
%if there exists a large number of such documents or a large number of processes to be analyzed \cite{leopold2018identifying}.

% \subsection{Experimental Setup and Dataset}

\subsection{Evaluation}

\begin{table}[b]
\vspace{-1em}
\centering
\caption{Exemplary Output of GPT4 for the RPA Classification Task}
\label{tab:ex_rpa}
\begin{tblr}{
  hlines,
  hline{1,8} = {-}{0.08em},
  hline{2-4} = {-}{0.05em},
}
\textbf{Task Input}                                              & \textbf{GPT4 Output} \\
register a claim performed by claims officer                  & User task      \\
examine a claim	performed by claims officer                   & Manual task   \\
write a settlement recommendation performed by claims officer & Manual task   \\
%4. check recommendation performed by claims officer              & (0 - Manual task)    \\
send the claim back to the claims performed by claims officer & User task
\\
%6. repeat the recommendation performed by claims officer         & (0 - Manual task)    
\end{tblr}
\end{table}

Following the approach from Fig.~\ref{fig:approach}, GPT4 is used to replicate the experiment of \cite{leopold2018identifying}. The task is described in the prompt by asking the LLM to classify process tasks into one of three output formats: manual, user, or automated task. Possible features that might affect the task classifications (e.g., verb feature, object feature, resource type (human/non-human), and IT domain) are included in the task description. The output format as well as an example of tasks' classification for tasks of a given process description is also provided. 
We use the same dataset as \cite{leopold2018identifying}, consisting of 33 textual process descriptions obtained from \cite{friedrich2}. These descriptions consist of 424 process tasks to be classified.
See Tab. \ref{tab:ex_rpa} for an example of an input process description and the output generated by GPT4.

% %\vspace{-2em}
% \begin{figure}[htb]
% \centering
% \caption{Exemplary textual process description and task classification from GPT4}
% \label{fig:rpa_example}
% \includegraphics[width=.8\textwidth]{figures/example_rpa.jpg}
% \end{figure} 
%\vspace{-1.5em}

%To evaluate input and output robustness, we retry the experiment five times. The first three tries aim at output robustness and use the same prompt formulated by one of the authors. Thus, they will report the variability of outputs and task classification given the same input.
%The remaining two tries use prompts written by two other authors and are applied to the same dataset to measure the input robustness of the GPT4 model on performing the same task using different inputs from different users. 
%The used prompts as well as the results returned by GPT4 can be found in the repository referenced in Sec.~\ref{sec:bpmn}. 
We did three identical prompts, we use two other prompts by different authors with an example in each. We also did one prompt without an example.
Table \ref{tab:eva_RPA } displays precision (Prec.), recall (Rec.), and F1-score (F1) for each of the six prompts compared to the benchmark from \cite{leopold2018identifying}.
For the overall results, we apply the same micro-averaging approach as the benchmark, i.e., the number of tasks belonging to a class was used to weigh the respective precision and recall values.

\begin{table}[tb]
\centering
\caption{Precision, Recall, and F1-Score for the RPA Task}
\label{tab:eva_RPA }

\begin{tabular}{|c|l|l||>{\centering\arraybackslash}m{1.6cm}>{\centering\arraybackslash}m{1.6cm}>{\centering\arraybackslash}m{1.7cm}|>{\centering\arraybackslash}m{1.4cm}|}
\hline
\multicolumn{1}{|l}{} & \multicolumn{1}{l}{}                                                        &                            & Manual~                    & User~ & Automated & Overall  \\ \hline \hline
\parbox[t]{7mm}{\multirow{9}{*}{\rotatebox[origin=c]{90}{\parbox[c]{2cm}{\centering Output Robustness}}}} & \multirow{3}{*}{\begin{tabular}[c]{@{}l@{}}Prompt 1\\ Try 1\end{tabular}}   & Prec.                      & 0.68                          & \textbf{0.88}
     & 0.15       & 0.79        \\
                      &                                                                             & Rec.                       & 0.83                         & 0.6     & 0.75        & 0.67        \\
                      &                                                                             & F1                         & 0.75                          & 0.74     & 0.45         & 0.73        \\ 
\cline{2-7}
                      & \multirow{3}{*}{\begin{tabular}[c]{@{}l@{}}Prompt 1\\ Try 2\end{tabular}}   & Prec.                      & 0.65                          & 0.84     & 0.73         & 0.78        \\
                      &                                                                             & Rec.                       & 0.69                         & 0.83     & 0.69         & 0.78        \\
                      &                                                                             & F1                         & 0.67                          & \textbf{0.84}     & \textbf{0.71}         & 0.78         \\ 
\cline{2-7}
                      & \multirow{3}{*}{\begin{tabular}[c]{@{}l@{}}Prompt 1\\ Try 3\end{tabular}}   & Prec.                      & 0.84                          & 0.84    & 0.32         & \textbf{0.82 }       \\
                      &                                                                             & Rec.                       & 0.65                          & 0.83     & \textbf{0.93}         & 0.78        \\
                      &                                                                             & F1                         & 0.75                         & \textbf{0.84}     & 0.63         & 0.8        \\ 
\hline
\parbox[t]{7mm}{\multirow{9}{*}{\rotatebox[origin=c]{90}{\parbox[c]{2cm}{\centering Input Robustness}}}} & \multirow{3}{*}{\begin{tabular}[c]{@{}l@{}} No Examples\end{tabular}} & Prec.
& \textbf{0.85}                          & 0.77     & 0.34        &    0.78     \\
                      &                                                                             & Rec.                       & 0.42                          & \textbf{0.88}     & 0.88         & 0.74         \\
                      &                                                                             & F1                         & 0.64                          & 0.83     & 0.61         &  0.76       \\  
\cline{2-7}
                      & \multirow{3}{*}{\begin{tabular}[c]{@{}l@{}}Other\\ Author (2)\end{tabular}} & Prec.                      & 0.44                          & 0.71     & 0.29         & 0.61         \\
                      &                                                                             & Rec.                       & 0.42                         & 0.74     & 0.13         &   0.62     \\
                      &                                                                             & F1                         & 0.43                          & 0.73     & 0.21        &  0.62      \\ 
                      \cline{2-7}
                      & \multirow{3}{*}{\begin{tabular}[c]{@{}l@{}}Other\\ Author (2)\end{tabular}} & Prec.                      & 0.5                          & 0.87     & 0.36         & 0.74        \\
                      &                                                                             & Rec.                       & 0.82                          & 0.55     & 0.8         & 0.64        \\
                      &                                                                             & F1                         & 0.66                          & 0.71     & 0.58         & 0.69        \\ 
\hline
\multicolumn{2}{|c|}{\multirow{3}{*}{Benchmark \cite{leopold2018identifying}}} & Prec. & 0.81 & 0.8 & \textbf{0.92} & 0.81 \\
\multicolumn{2}{|c|}{} & Rec. & \textbf{0.9} & 0.7 & 0.52 & \textbf{0.8} \\
\multicolumn{2}{|c|}{} & F1 & \textbf{0.85} & 0.75 & 0.66 & \textbf{0.81} \\ \hline 
\end{tabular}
%\vspace{-4em}
\end{table}

GPT4 outperforms the benchmark for 4 out of 6 prompts for user tasks. For the automated tasks, precision results are below the benchmark because many tasks were classified by GPT4 as automated although they are not. However, the recall for this class outperforms the benchmark in almost all the prompts. For the F1-score, it is similar to the benchmark for the classes except for the user class where the F1-score results were higher than the benchmark. Overall, as indicated by the F1-score, GPT4 performs similarly to the benchmark for all six prompts. 
We also saw the performance of GPT4 deteriorate over time.
%The classification for tasks that are provided at the beginning of the prompts is better than tasks that are provided later. 
We suspect that this is caused by the limited context window of GPT4, combined with the large number of tasks to be classified (424). In such cases, reminding the LLM of the task description between inputs could yield better results.

% \subsection{Discussion}
% We can see that results from GPT4 model performed better in the task classification tasks since clear instructions are given for the classification. However, the model did not produce satisfactory results for the task identification part which affected the evaluation results since a large number of classifications are not performed since not all the tasks are identified. To solve this problem, we recommend either giving the tasks for the model ready for classification or provide clear instructions on how process tasks can be identified from textual process descriptions.

\section{Discussion} \label{sec:disc}

After illustrating that out-of-the-box GPT4 performs similarly or even better than specialized approaches for our three exemplary tasks, we now want to discuss the usage of LLMs in practice and provide guidelines for users.
%and implications for future research. 
%In particular, we want to discuss the evaluation necessities due to the non-deterministic character of LLMs and the usage implications if practitioners apply our approach.

%\subsection{Usage of LLMs in Practice}

\mypar{Prompt Recommendations} 
In our experiments, we found that including different contents in the prompt increase the performance of GPT4. For example, the output should be clearly defined instigating the task. Further, for the text-to-LTL task, examples led to better results. We can therefore recommend specifying the output format and to try using examples if feasible. In general, different prompts should be used and compared to maximize the benefits of using GPT4.

\mypar{Non-deterministic output} 
In order to produce more natural-sounding text, generative LLMs typically have \emph{temperature} parameter that adds some variability to the output. Because of this, responses given by GPT4 may change even if the input remains constant. At the same time, if the input is varied slightly (e.g., by phrasing the same instruction in a different way), the model may make significant alterations to its response.
In our experiments, we attempted to account for this by establishing a certain level of input and output consistency.
%the former by employing prompts written by three different authors, and the latter by repeatedly evaluating the results produced by the same prompt. 
We found that, although results are overall relatively consistent, there is still considerable variation in how well each response reflects individual aspects of the provided text, for example, whether a particular task has been correctly identified and categorized.
We, therefore, argue that future research into the behavior of LLMs and their reaction to different inputs is needed. 
%This could include prompt engineering studies in which prompts are generated by different groups of authors (for example, with varying levels of expertise), or studies that compare the output of different LLMs. We also believe that it is crucial to consider the variability in output when using these models in BPM research, especially when the objective is to translate text to a formal representation like a Declare constraint. 
In particular, the non-deterministic nature of LLM's output has implications for evaluation design: in our opinion, a basic sensitivity analysis as applied in this paper is always required in order to perform a meaningful evaluation of performance.

% guideline 2: provide target template
% guideline 3: consider providing examples
% other guidelines?

\mypar{File Generation} 
%\mypar{Use Case Application} 
When using it in practice, as illustrated with the three tasks, GPT4 does not generate files but rather text. Therefore, in order to use it in the first two exemplary tasks, further translation into formalized languages was necessary. This can be done via a compiler that generates \emph{Declare} constraints or BPMN models based on the output. Nevertheless, it poses a limitation of current LLMs, especially considering output variability. It should be noted that this limitation is specific to present-day LLMs such as GPT4, which are not capable of file generation and may be overcome by future iterations of the models.

\section{Conclusion} \label{sec:concl} 
In this paper, we developed and applied an approach that utilizes the LLM GPT4 for diverse BPM tasks. The approach itself is simple and leverages the capabilities of GPT4 by instructing it to accomplish the task at hand. We selected three BPM tasks to illustrate that GPT4 is indeed able to accomplish them: mining imperative process models from the textual description, mining declarative process models from the textual description, and assessing RPA suitability of process tasks from textual descriptions. For all three tasks, GPT4 performs similarly to or better than the benchmark, i.e., specific applications for the respective task. We analyzed the input and output robustness of the approach and found that the output is relatively insensitive to different executions of the same prompt, even if different authors formulated them. Further, we found that some prompts should include examples to help the LLM.
Future research could assess whether LLMs are also applicable to other tasks from different phases of the BPM lifecycle.
All in all, this paper illustrates and evaluates three practical applications of GPT4 and provides 
%theoretical and practical 
implications for future research and usage. 

\newpage

\bibliographystyle{splncs04}
\bibliography{bibliography}

\begin{thebibliography}{10}
\providecommand{\url}[1]{\texttt{#1}}
\providecommand{\urlprefix}{URL }
\providecommand{\doi}[1]{https://doi.org/#1}

\bibitem{PET}
Bellan, P., van~der Aa, H., Dragoni, M., Ghidini, C., Ponzetto, S.P.: Pet: An
  annotated dataset for process extraction from natural language text tasks.
  In: BPM Workshops. pp. 315--321. Springer (2023)

\bibitem{Bellan2020}
Bellan, P., Dragoni, M., Ghidini, C.: A qualitative analysis of the state of
  the art in process extraction from text. In: DP@AI*IA (2020)

\bibitem{Bellan2022}
Bellan, P., Dragoni, M., Ghidini, C.: Extracting business process entities
  and relations from text using pre-trained language models and in-context
  learning. In: Enterprise Design, Operations, and Computing. pp. 182--199.
  Springer (2022)

\bibitem{burattin2016conformance}
Burattin, A., Maggi, F.M., Sperduti, A.: Conformance checking based on
  multi-perspective declarative process models. Expert Syst Appl  \textbf{65},
  194--211 (2016)

\bibitem{busch2023just}
Busch, K., Rochlitzer, A., Sola, D., Leopold, H.: Just tell me: Prompt
  engineering in business process management. preprint arXiv:2304.07183  (2023)

\bibitem{bpm_book}
Dumas, M., La~Rosa, M., Mendling, J., Reijers, H.: {Introduction to Business
  Process Management}. In: Fundamentals of Business Process Management.
  Springer (2018)

\bibitem{friedrich2}
Friedrich, F.: {Automated generation of business process models from natural
  language input (Master thesis)}, \url{https://frapu.de/pdf/friedrich2010.pdf}

\bibitem{friedrich2011process}
Friedrich, F., Mendling, J., Puhlmann, F.: {Process Model Generation from
  Natural Language Text}. In: CAiSE. pp. 482--496. Springer (2011)

\bibitem{klievtsova2023conversational}
Klievtsova, N., Benzin, J.V., Kampik, T., Mangler, J., Rinderle-Ma, S.:
  Conversational process modelling: State of the art, applications, and
  implications in practice. preprint arXiv:2304.11065  (2023)

\bibitem{leopold2018identifying}
Leopold, H., {van der Aa}, H., Reijers, H.A.: Identifying candidate tasks for
  robotic process automation in textual process descriptions. In: BPMDS. pp.
  67--81. Springer (2018)

\bibitem{mustansir2022towards}
Mustansir, A., Shahzad, K., Malik, M.K.: Towards automatic business process
  redesign: an {NLP} based approach to extract redesign suggestions. Autom
  Softw Eng  \textbf{29},  1--24 (2022)

\bibitem{openai2023gpt4}
OpenAI: {GPT-4 Technical Report}. preprint arXiv:2304.04309  (2023)

\bibitem{rebmann2021extracting}
Rebmann, A., {van der Aa}, H.: Extracting semantic process information from the
  natural language in event logs. In: CAiSE. pp. 57--74. Springer (2021)

\bibitem{reddy2019study}
Reddy, K.N., Harichandana, U., Alekhya, T., Rajesh, S.: A study of robotic
  process automation among artificial intelligence. Int J Sci Res
  \textbf{9}(2),  392--397 (2019)

\bibitem{reijers2003product}
Reijers, H.A., Limam, S., {van der Aalst}, W.: {Product-Based Workflow Design}.
  JMIS  \textbf{20}(1),  229--262 (2003)

\bibitem{rizun2021assessing}
Rizun, N., Revina, A., Meister, V.G.: Assessing business process complexity
  based on textual data: Evidence from {ITIL} {IT} ticket processing. BPMJ
  \textbf{27}(7),  1966--1998 (2021)

\bibitem{teubner2023welcome}
Teubner, T., Flath, C.M., Weinhardt, C., {van der Aalst}, W., Hinz, O.: Welcome
  to the era of {ChatGPT} et al. {The} prospects of large language models. BISE
   \textbf{65},  95--101 (2023)

\bibitem{Aa2018}
{van der Aa}, H., Carmona, J., Leopold, H., Mendling, J., Padr{\'o}, L.:
  {Challenges and Opportunities of Applying Natural Language Processing in
  Business Process Management}. In: COLING. pp. 2791--2801 (2018)

\bibitem{van2019extracting}
{van der Aa}, H., Di~Ciccio, C., Leopold, H., Reijers, H.A.: Extracting
  declarative process models from natural language. In: CAiSE. pp. 365--382.
  Springer (2019)

\bibitem{vidgof2023large}
Vidgof, M., Bachhofner, S., Mendling, J.: Large language models for business
  process management: Opportunities and challenges. preprint arXiv:2304.04309
  (2023)

\end{thebibliography}

\end{document}